\def\year{2021}\relax
\documentclass[letterpaper]{article} 
\usepackage{aaai21}  

\usepackage{times}  
\usepackage{helvet} 
\usepackage{courier}  
\usepackage[hyphens]{url}  
\usepackage{graphicx} 
\def\UrlFont{\rm}  
\usepackage{natbib}  
\usepackage{caption} 
\usepackage{preamble}
\usepackage{tikz}
\usetikzlibrary{graphs}
\usetikzlibrary{graphs.standard}
\usetikzlibrary{shapes, arrows}
\usetikzlibrary{calc,arrows,decorations.pathmorphing,backgrounds,fit,positioning,shapes.symbols,chains}
\usetikzlibrary{shapes.geometric}
\usetikzlibrary{decorations.pathreplacing}
\usepackage[normalem]{ulem}
\usepackage{url}
\usepackage[switch]{lineno}
\usepackage{cancel}
\usepackage{algorithm, algorithmic}

\newcommand{\paragraf}[1]{\noindent\textbf{#1}\hspace{4pt}}

\DeclareMathOperator{\bern}{Bernoulli}

\DeclareMathOperator{\ig}{IG}
\DeclareMathOperator{\lognormal}{Lognormal}
\DeclareMathOperator{\multi}{multi}
\DeclareMathOperator{\single}{single}

\counterwithout{equation}{section}

\title{Learning Compositional Sparse Gaussian Processes with a Shrinkage Prior}
\author{
Anh Tong\textsuperscript{\rm 1},
Toan Tran\textsuperscript{\rm 2},
Hung Bui\textsuperscript{\rm 2},
Jaesik Choi\textsuperscript{\rm 3,4}\\
}

\affiliations{
    \textsuperscript{\rm 1} Ulsan National Institute of Science and Technology\\
    \textsuperscript{\rm 2} VinAI Research\\
    \textsuperscript{\rm 3} Korea Advanced Institute of Science and Technology\\
    \textsuperscript{\rm 4} INEEJI
    \\anhth@unist.ac.kr, \{v.toantm3,v.hungbh1\}@vinai.io, jaesik.choi@kaist.ac.kr
}

\begin{document}
\maketitle
\begin{abstract}
    Choosing a proper set of kernel functions is an important problem in learning Gaussian Process (GP) models since each kernel structure has different model complexity and data fitness. Recently, automatic kernel composition methods provide not only accurate prediction but also attractive interpretability through search-based methods. {However, existing methods suffer from slow kernel composition learning.} To tackle large-scaled data, we propose a new sparse approximate posterior for GPs, MultiSVGP, constructed from groups of inducing points associated with individual additive kernels in compositional kernels. We demonstrate that this approximation provides a better fit to learn compositional kernels given empirical observations. We also provide theoretically justification on error bound when compared to the traditional sparse GP. In contrast to the search-based approach, we present a novel probabilistic algorithm to learn a kernel composition by handling the sparsity in the kernel selection with Horseshoe prior. We demonstrate that our model can capture characteristics of time series with significant reductions in computational time and have competitive regression performance on real-world data sets.
\end{abstract}

\section{Introduction}
Recently, there are numerous advancements in automating model learning with statistical methods. There are still many challenging problems including making the automatic procedure efficient and applying methods on large-scaled data.

The Automatic Statistician framework~\cite{grosse2012a,DuvLloGroetal13,Lloyd2014ABCD,Ghahramani15_nature,abcd_book} aims to address challenges on automating model discovery. The framework adopts search procedures over a space of models, enumerating possible compositional Gaussian Process (GP) kernel functions generated from base ones and compositional grammar rules. Learned models can produce human-readable explanations as dissecting small explainable components in the compositional kernel models. However, existing methods suffer from slow kernel composition learning due to the huge search space. 

Recently, a differential compositional learning inspired by deep neural networks ~\cite{differential_ckl} is proposed. Although this model obtains expressive kernel functions for GPs and consequently achieves good predictive performances, it is hard to interpret compared to existing Bayesian kernel learning methods, e.g. ~\cite{Lloyd2014ABCD}.

This paper presents a new kernel composition learning method which encourages to learn a sparse composition of kernels with a shrinkage prior, Horseshoe prior,  which is proven to be effective in learning sparse signal~\cite{horseshoe,lasso_horseshoe}.
To retain the model interpretability, we formulate compositional kernels as the sum of individual additive kernels which can be explained in natural language.

To scale up the model for large-scaled data, we introduce a new approximate posterior for GP, Multi-inducing Sparse Variational GP ({MultiSVGP}). Existing sparse inducing GP~\cite{snelson_2006,sgp_bigdata} methods give an approximation for GP for a kernel function in general with a single set of inducing points. However, we can further improve this approach specifically for additive compositional kernels by considering a group of inducing points and assigning an individual member of inducing points responsible for an additive kernel in our approximating posterior. We justify that our approximation with compositional kernels produces a better error bound than the sparse inducing GP. In experiments, we demonstrate that our models can capture appropriate kernel structures for time series from small-scaled data sets to large-scaled data sets. In general, our model runs faster than existing compositional kernel learning methods~\cite{Lloyd2014ABCD, scale_abcd_kim2018a} five to twenty times while maintaining similar accuracy. We also show that our model to have competitive extrapolation performance in regression tasks as well as improve additive GPs~\cite{additive_gp} with our kernel selection. 

The paper offers the following contributions: (1) a new accurate GP approximation for additive compositional kernels; (2)  kernel selection using Horseshoe prior so that the
learned models can capture the inductive kernel structure in data and remain interpretable.

\section{Related work}
There is a large body of work~\cite{DuvLloGroetal13,Lloyd2014ABCD,scale_abcd_kim2018a} establishing the foundation of model discovery for Gaussian processes. \cite{grosse2012a} presents work on unsupervised learning for the case of matrix decomposition. Then,~\cite{DuvLloGroetal13,Lloyd2014ABCD,Ghahramani15_nature,Hwangb16,tong19a} extend to supervised settings with Gaussian process (GP) models. \cite{differential_ckl} build complex kernels under network architectures having an additive layer where any two kernels are summed and followed by a product layer where kernels are multiplied. \cite{scale_abcd_kim2018a} adopt search procedure but smartly avoid the learning model by finding bounds of likelihood. However, the search is still time-consuming and is done heuristically greedy manner. A complete review as well as a guideline for automatic systems can be found in~\cite{abcd_book}. While inheriting the spirits of existing work on kernel compositions, this paper focuses on scaling up the system in terms of data size and efficient model selection. 

A recent work~\cite{aaai_kernel_selection_svgp} shares some similarity to our work. The paper presents a probabilistic approach to select among models with a Softmax-like assumption for choosing a model. The main idea in this paper is to select a single model out of a manual fixed set of candidate models. Our model generating kernels combinatorially considers a bigger model space than that of~\cite{aaai_kernel_selection_svgp}. On the other hand, models~\cite{aaai_kernel_selection_svgp} use single inducing points for compositional kernels. Its sparse GP approximation can be limited compared to our MultiSVGP. Another work~\cite{BO_abcd} attempts to extract a model out of candidate ones using surrogate models, e.g., Bayesian optimization. The surrogate models are based on the distance between GP models. A recent approach on learning implicit kernel~\cite{bbq} seeks for the representation in their corresponding spectral domain. Exploiting the spectral representation is previously used in~\cite{WilsonA13}. However, the question of how to make these methods interpretable left unanswered. 

There is existing work proposing probabilistic priors, {e.g.,} spike-and-slab prior, on multi-task GP~\cite{multi_task_spike_and_slab}. However, the approach does not scale {well} with the number of data and suffers from computational burden by the choice of probabilistic priors. Other work that is related but distinct includes~\cite{automporphing,nonstationary_covariance,abcd_pp,bayes_synth}.

\section{Background}
This section presents reviews on building blocks of this paper including Gaussian process, the kernel selection framework, and shrinkage prior.
\subsection{Gaussian Process}
\acrfull{gp} is known as a nonparametric prior over function values~\cite{Rasmussen_GPM}. Formally, a~\acrshort{gp} is defined as a multivariate Gaussian distribution over function value $f(\vx)$:
$$f(\vx) \sim \GP(m(\vx), k(\vx,\vx')),$$
where $m(\vx)$ is the mean function (usually is set as zero mean), and $k(\vx,\vx')$ is the kernel (covariance) function. Given data $\mathcal{D}=(\vX,\vy)$, the predictive posterior distribution at $\vx_*$ is in a closed form:
\begin{align*}
    f(\vx_*)| \vX,\vy &\sim \Normal(\mu(\vx_*), \sigma^2(\vx_*)),\\
    \mu(\vx_*) &= k(\vx_*, \vX)k(\vX, \vX)^{-1}\vy,\\
    \sigma^2(\vx_{*}) &= k(\vx_*, \vx_*) - k(\vx_*, \vX)k(\vX, \vX)^{-1}k(\vx_*, \vX)^\top,
\end{align*}
where $k(\vx_*, \vX)$ computes the covariance between function evaluations at $\vx_*$ and $\vX$. This property of~\acrshort{gp} provides an efficient way to calibrate model uncertainties, becoming useful in several calibration methods, i.e. Bayesian optimization and Bayesian quadrature.
\subsection{The Automatic Statistician}
The Automatic Statistician or~\acrfull{abcd} aims to mimic and automate the process of statistical modeling~\cite{grosse2012a,DuvLloGroetal13,Lloyd2014ABCD,Ghahramani15_nature}. There are three main components in this framework: language of models, search procedure and report generations.\\
\paragraf{Language of models}\gp~models are constructed by a grammar over kernels with a set of base kernels and kernel operators. The base kernels are: \kSE~(squared exponential), \kLin~(linear),~\kPer~(periodic) (see Appendix~\ref{appendix:kernels} for details). The operators consist of $+$ (addition), $\times$ (multiplication). As composed kernels get more complex, the corresponding generated models become more expressive to fit complex data.\\
\paragraf{Search procedure} The search procedure is done in a greedy manner. That is, the language of models generates candidate models. Then, all of them are optimized by maximizing log likelihoods. A model is selected based on the trade-off between model and data complexity. This selected model is the input of the language of models to create new candidates.\\
\paragraf{Automatic generated explanation of models} The compositional kernels resulted from the search procedure are transformed into a sum of products of base kernels. Each structural product of kernels is interpreted under natural-language expressions. All descriptions are gathered to produce a complete report with visualized plots and human-friendly analyses.

\subsection{Sparse Variational Gaussian Process}
The history of sparse Gaussian process methods dated back from the work of~\cite{snelson_2006}.  It can be considered as a sibling of Nystr\"{o}m approximation~\cite{nystrom_william_2000}. The main idea of sparse Gaussian process is to introduce \emph{pseudo inducing points}, $\vu$, which are distributed jointly with Gaussian process latent variable $\vf$ under a Gaussian distribution. The number of inducing points, $M$, is much smaller than the number of data points, $N$. The computational complexity of sparse Gaussian process is $\mathcal{O}(NM^2)$ for each learning iteration. There is a line of research on improving and understanding sparse Gaussian processes~\cite{fitc_vfe_2016,BuiYT17,rate_convergence_burt19a}. 

Given a data set $\mathcal{D} = \{(\vx_i, y_i)\}_{i=1}^n$, a sparse variational Gaussian process (SVGP) is defined by
\begin{align*}
    f(\cdot) & \sim \mathcal{GP}(0, k(\cdot, \cdot)) , \\
    y_i | f, \vx_i & \sim p(y_i| f(\vx_i)),
\end{align*}
where $p(y_i|f(\vx_i))$ is a likelihood, e.g., Gaussian, categorical. To approximate the posterior, the variational approach considers $M$ inducing points $\vu = \{u_i\}_{i=1}^M$ at locations $\{\vz_i\}_{i=1}^M$, forming the variational distribution as
\begin{align*}
    \vu &\sim \Normal(\vm, \vS), \\
    f(\cdot)|&\vu \sim \mathcal{GP}(\mu(\cdot), \Sigma(\cdot, \cdot)), 
\end{align*}
where the mean and and covariance are obtained as 
\begin{equation}
\begin{aligned}
    \mu(\cdot) & = \vk_\vu(\cdot)^\top \vK_{\vu\vu}^{-1}\vu,\\
    \Sigma(\cdot, \cdot) & = k(\cdot, \cdot)  - \vk_\vu(\cdot)^\top\vK_{\vu\vu}^{-1}\vk_\vu(\cdot),
\end{aligned}
\label{eq:mean_var_sgp}
\end{equation}
with $\vk_\vu(\cdot) = [k(\vz_i, \cdot)]_{i=1}^M$ and $\vK_{\vu\vu}$ is the covariance of $\vu$.

The variational inference maximizes the evidence lower bound (ELBO) given as following~\cite{sgp_bigdata}
\begin{equation*}
    \mathcal{L} = \sum_i\expect_{q(f(\cdot))}\left[\log p(y_i|f(\vx_i))\right] - \KL[q(\vu)||p(\vu)].
\end{equation*}
Here $q(f(\cdot))$ is a Gaussian, having mean as $\vk_\vu(\cdot)^\top \vK_{\vu\vu}^{-1}\vm$ and covariance as $ k(\cdot, \cdot)  - \vk_\vu(\cdot)^\top \vK_{\vu\vu}^{-1} (\vS - \vK_{\vu\vu})\vK_{\vu\vu}^{-1}\vk_\vu(\cdot)$.
The objective $\mathcal L$ can be optimized using stochastic gradient descent. 

\subsection{Shrinkage prior}
In many model learning scenarios, we usually face the problem of sparse modeling for variable selection. Some prominent methods are proposed to tackle the problem, including Lasso regularization~\cite{lasso_paper}, spike and slab prior~\cite{spike_and_slab_paper} and Horseshoe prior~\cite{horseshoe}.
\paragraph{Spike and slab prior} The spike and slab prior over $\vw$ is defined as
\begin{align*}
    v_i &\sim \Normal(0, \sigma^2_w),\\
    s_i & \sim \bern(\pi_s),\\
    w_i & = v_is_i.
\end{align*}
This belongs to the two-group models. That is, the event $\{w_i = 0\}$ happens with probability $(1-\pi_s)$; and nonzero $w_i$ distributed according to a Gaussian prior with probability $\pi_s$. There is existing work using the prior for kernel learning~\cite{multi_task_spike_and_slab}. 
\paragraph{Horseshoe prior}
The horseshoe prior~\cite{horseshoe} introduces a way to sample a sparse vector $\boldsymbol{\beta}$ as
\begin{align*}
    \beta_i & \sim \Normal(0, \tau^2 \lambda^2_i), \quad i=1\dots m \\
    \lambda_i & \sim \mathcal{C}^+(B), \quad i=1\dots m\\
    \tau &\sim \mathcal{C}^+(A),
\end{align*}
where $\mathcal{C}^{+}(\cdot)$ is the half-Cauchy distribution, $A$ and $B$ are the scale parameters. Here, $\tau$ is the global shrinkage parameter, $\lambda_i$ is the local shrinkage parameter. In contrast to the spike and slab prior, horseshoe prior is a continuous shrinkage {one}. It has Cauchy-like tails which allow signals to at large values. On the other hand, the infinite spike near zero keeps $w_i$ around the origin.~\cite{horseshoe_bnn} uses Horseshoe prior for the weight selection in Bayesian deep neural networks.

Compared to Horseshoe prior, the spike and slab prior exhibits a substantial computational burden as the dimension of sparse vectors increases.

\def\r{2.4}
\def\rn{3.4}
\def\n{2} \def\myangles{{175,90,5}}
\def\myanglesinducing{{165,90,15}}
\def\m{1} \def\inducingangles{{240, 300}}
\newcounter{np} \pgfmathsetcounter{np}{\n+1}
\newcounter{na} \newcounter{nb} \newcounter{nc}
\newcounter{ia} 
\pgfmathsetcounter{na}{\n-1}    
\pgfmathsetcounter{nb}{\n-2}    
\pgfmathsetcounter{nc}{\n-3}    
\newcounter{q} \setcounter{q}{0}    
\newcounter{e} \setcounter{e}{0}    
\newcounter{a} \setcounter{a}{0}    
\newcounter{b} \setcounter{b}{1}    
\newcounter{c} \setcounter{c}{2}    
\newcounter{d} \setcounter{d}{2}    

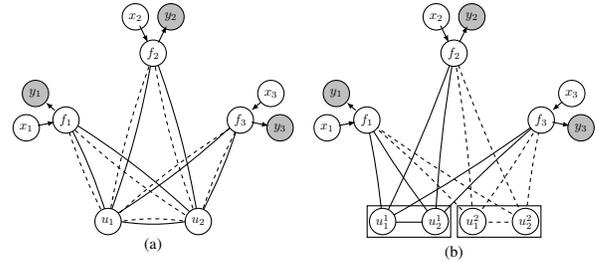
\begin{figure}
\centering
\scalebox{0.5}{%
\begin{tikzpicture}

        
                


\begin{scope}[shift={(8,2)}]

\foreach \i in {0,...,\m}{
    \pgfmathparse{\inducingangles[\i]} \let\t\pgfmathresult
    \foreach \j in {0,...,\n}{
        \pgfmathparse{\myanglesinducing[\j]} \let\u\pgfmathresult
        \draw[thick] ({\r*cos(\t)},{\r*sin(\t)+0.05}) to[bend right=5] ({\r*cos(\u)},{\r*sin(\u)+0.05}) ;
            \draw[thick, dashed] ({\r*cos(\t)},{\r*sin(\t)-0.05}) to[bend left=5] ({\r*cos(\u)},{\r*sin(\u) -0.05});
    }
}

\pgfmathparse{\inducingangles[0]} \let\t\pgfmathresult
\pgfmathparse{\inducingangles[1]} \let\u\pgfmathresult
\draw[thick] ({\r*cos(\t)},{\r*sin(\t)+0.05}) to[bend right=10] ({\r*cos(\u)},{\r*sin(\u)+0.05}) ;
\draw[thick, dashed] ({\r*cos(\t)},{\r*sin(\t)-0.05}) to[bend left=10] ({\r*cos(\u)},{\r*sin(\u) -0.05});

\foreach \i in {0,...,\m}{
    \pgfmathparse{\inducingangles[\i]} \let\t\pgfmathresult
    \pgfmathsetcounter{ia}{\i+1}
    \fill[draw=black,fill=white!20!,thick] ({\r*cos(\t)},{\r*sin(\t)})circle (3.5mm) node (f_\theia) {$u_\theia$};
}

\foreach \i in {0,...,\n}{
        \pgfmathparse{\myanglesinducing[\i]}    \let\t\pgfmathresult
        \pgfmathsetcounter{ia}{\i+1}
        \fill[draw=black,fill=white!20!,thick]
                ({\r*cos(\t)},{\r*sin(\t)})circle (3.5mm) node (f_\theia) {$f_\theia$};
        \fill[draw=black,fill=white!20!,thick]
                ({\rn*cos(\t +8)},{\rn*sin(\t + 8)})circle (3.5mm) node (x_\theia) {$x_\theia$};
        
        \fill[draw=black,fill=gray!50!,thick]
                ({\rn*cos(\t - 8)},{\rn*sin(\t - 8)})circle (3.5mm) node (y_\theia) {$y_\theia$};
                
         \draw[draw, -latex, thick] (x_\theia)--(f_\theia);
         \draw[draw, -latex, thick] (f_\theia)--(y_\theia);
}
\node at (0,-2.7) (inducing_gp) {\Large{(a)}};

\end{scope}

\begin{scope}[shift={(16,2)}]

\pgfmathparse{\inducingangles[0]} \let\t\pgfmathresult
\foreach \j in {0,...,\n}{
    \pgfmathparse{\myanglesinducing[\j]} \let\u\pgfmathresult
    \draw[thick] ({\r*cos(\t)-0.7},{\r*sin(\t)}) to[bend right=3] ({\r*cos(\u)},{\r*sin(\u)}) ;
     \draw[thick] ({\r*cos(\t) + 0.7},{\r*sin(\t)}) to[bend left=3] ({\r*cos(\u)},{\r*sin(\u)});
}
\draw[thick] ({\r*cos(\t)-0.7},{\r*sin(\t)}) to[bend right=3] ({\r*cos(\t) + 0.7},{\r*sin(\t)}) ;

\pgfmathparse{\inducingangles[1]} \let\t\pgfmathresult
\foreach \j in {0,...,\n}{
    \pgfmathparse{\myanglesinducing[\j]} \let\u\pgfmathresult
    \draw[thick, dashed] ({\r*cos(\t)-0.7},{\r*sin(\t)}) to[bend right=3] ({\r*cos(\u)},{\r*sin(\u)}) ;
     \draw[thick, dashed] ({\r*cos(\t) + 0.7},{\r*sin(\t)}) to[bend left=3] ({\r*cos(\u)},{\r*sin(\u)});
}
\draw[thick, dashed] ({\r*cos(\t)-0.7},{\r*sin(\t)}) to[bend right=3] ({\r*cos(\t) + 0.7},{\r*sin(\t)}) ;

\foreach \i in {0,...,\m}{
    \pgfmathparse{\inducingangles[\i]} \let\t\pgfmathresult
    \pgfmathsetcounter{ia}{\i+1}
    \fill[draw=black,fill=white!20!,thick] ({\r*cos(\t)-0.7},{\r*sin(\t)})circle (3.5mm) node (f_\theia_1) {$u^\theia_1$};
    \fill[draw=black,fill=white!20!,thick] ({\r*cos(\t)+0.7},{\r*sin(\t)})circle (3.5mm) node (f_\theia_2) {$u^\theia_2$};
    \draw[thick] ($(f_\theia_1.south west) + (-0.1, -0.1)$) rectangle ($(f_\theia_2.north east) + (0.1, 0.1)$);
}

\foreach \i in {0,...,\n}{
        \pgfmathparse{\myanglesinducing[\i]}    \let\t\pgfmathresult
        \pgfmathsetcounter{ia}{\i+1}
        \fill[draw=black,fill=white!20!,thick]
                ({\r*cos(\t)},{\r*sin(\t)})circle (3.5mm) node (f_\theia) {$f_\theia$};
        \fill[draw=black,fill=white!20!,thick]
                ({\rn*cos(\t +8)},{\rn*sin(\t + 8)})circle (3.5mm) node (x_\theia) {$x_\theia$};
        
        \fill[draw=black,fill=gray!50!,thick]
                ({\rn*cos(\t - 8)},{\rn*sin(\t - 8)})circle (3.5mm) node (y_\theia) {$y_\theia$};
                
         \draw[draw, -latex, thick] (x_\theia)--(f_\theia);
         \draw[draw, -latex, thick] (f_\theia)--(y_\theia);
}

\node at (0,-2.9) (structure_inducing_gp) {\Large{(b)}};
\end{scope}
\end{tikzpicture}
}
\caption{
The graphical model of two models. Solid and dashed lines indicate the connections modeled by two different kernel function $k_1$ and $k_2$. (a): Sparse inducing GP. The inducing points $u_i$ is introduced as a proxy for the connections between $f_i$. (b): Our approach. Inducing points are grouped. Each group represents an individual kernel $k_1$ or $k_2 $.}
\label{fig:model}
\end{figure}

\section{Kernel selection with shrinkage prior}
This section presents our main contributions: (1) kernel selection with Horseshoe prior and (2) our approximate GP for compositional kernels.

\subsection{Kernel selection with Horseshoe prior}
 Consider the full GP model with kernel construction based on generative procedures:
\begin{align}
    f(\cdot)|\vw  \sim \mathcal{GP}(0, \tilde{k}(\cdot, \cdot)),
    \label{eq:full_gp_horseshoe}
\end{align}
where $\tilde{k}(\vx, \vx') = \sum_{i=1}^m w_i^2 k_i(\vx, \vx')$ is constructed from $m$ kernel functions $k_i(\vx, \vx')$. We introduce a probabilistic prior over the weights $\vw=[w_{1:m}]$, $p(\vw)$ motivated from the Horseshoe prior. That is, we add the covariance term $k_i(\vx, \vx')$ to the step sampling $\beta_i$ in Horseshoe generative procedure. This makes $\beta_i$ equivalent to $f_i(\vx)$ in GP, i.e.,
\begin{equation*}
    \beta_i \sim \Normal(0, \tau^2\lambda_i^2) \Rightarrow f_i(\vx) \sim \mathcal{GP} (0, \tau^2 \lambda^2_i k_i(\vx, \vx')).
\end{equation*}
When considering the multivariate normal distribution $\vf_i \sim \Normal(\mathbf{0}, \tau^2\lambda_i^2\vK_i)$ with kernel matrix $\vK_i$ computed from $k_i(\vx, \vx')$, the \emph{multivariate} version of Horseshoe variable, $\boldsymbol{\beta}_i \sim \Normal(\mathbf{0}, \tau^2\lambda_i^2 \mathbf{I})$, is a special case of $\vf_i$ when $\vK_i$ is the identity matrix. This generalization is natural, equipping the sparsity among $\{f_i(\vx)\}_{i=1}^m$. Denoting $w_i^2 = \tau^2\lambda_i^2$ and assuming that $\{f_i(\vx)\}_{i=1}^m$ are mutually independent, we can get $f(\vx) = \sum_{i=1}^m f_i(\vx) \sim \mathcal{GP}(0, \tilde{k}(\vx, \vx'))$. 

The assumption on sparsity among kernel functions $k_i$ encourages simple kernels which agree with model selection principles like Occam's razor in~\cite{occams_zaros} and BIC in~\cite{Lloyd2014ABCD}.  

\subsection{Multi-inducing sparse Gaussian process}
To motivate the proposed approach, we will first provide a naive model directly obtained from the sparse Gaussian process. Then, we present our main model.\\
\paragraf{SVGP with compositional kernels}
Given inducing points $\vu$, we formulate the corresponding sparse GP model as 
\begin{align}
    f(\cdot)|\vw, \vu & \sim \mathcal{GP}(\tilde{\mu}(\cdot), \tilde{\Sigma}(\cdot, \cdot)), \label{eq:naive_model}\\ 
    \tilde{\mu}(\cdot) & = \tilde{\vk}_\vu^\top(\cdot) \tilde{\vK}_{\vu\vu}^{-1}\vu, \nonumber\\
    \tilde{\Sigma}(\cdot, \cdot) &= \tilde{k}(\cdot, \cdot) - \tilde{\vk}_\vu^\top(\cdot) \tilde{\vK}_{\vu\vu}^{-1} \tilde{\vk}_\vu(\cdot). \nonumber
\end{align}
Here, we denote that $\tilde{\vK}_{\vu\vu} = \sum_{i=1}^m w_i^2\vK_{i\vu\vu}$, and $\vK_{i\vu\vu}$ is the covariance of $\vu$ computed from kernel function $k_i(\cdot, \cdot)$.\\
\paragraf{Multi-inducing sparse variational GP} Given $\vw$, we define the model via the combination of conditional posterior distributions:
\begin{align}
    f(\cdot)|\vU, \vw \sim \mathcal{GP}\left(\sum_{i=1}^m w_i \mu_i(\cdot;\vu_i), \sum_{i=1}^mw_i^2 \Sigma_i(\cdot, \cdot;\vu_i)\right), 
    \label{eq:decompose_conditional}
\end{align}
where
\begin{equation}
    \begin{aligned}
        \mu_i(\cdot) &= \vk_{\vu_i}(\cdot)^\top \vK_{\vu_i\vu_i}^{-1}\vu_i,\\ 
        \Sigma_i(\cdot, \cdot) & = k_i(\cdot, \cdot) - \vk_{\vu_i}(\cdot)^\top \vK_{\vu_i\vu_i}^{-1}\vk_{\vu_i}(\cdot).
    \end{aligned}
    \label{eq:mu_sigma}
\end{equation}
Here $k_{\vu_i}(\cdot) = [k_i(\vu_i, \cdot)]^\top$, and $\vK_{\vu_i\vu_i}$ is the covariance of $\vu_i$ w.r.t. kernel $k_i$. For convenience, we omit the notation $\vu_i$ in $\mu_i$ and $\Sigma_i$. 

In contrast to the model in Equation~\ref{eq:naive_model}, we use $m$ inducing groups of inducing points, $\vU = \{\vu_i\}_{i=1}^m$. Each $\vu_i$ is associated with a kernel function $k_i$, consisting of $M_i$ inducing points $\vu_i = [u^{(i)}_{1:M_i}]$ at inducing location $\vZ_i = [\vz^{(i)}_{1:M_i}] \in \mathcal{Z}_i$. The number of inducing points $M_i$ is chosen to be much smaller than the size of data set $N$, i.e. ($M_i \ll N$).

Figure~\ref{fig:model} is an illustrative comparison between SVGP and our proposed approach. In plain words, each member in inducing groups is responsible for a single kernel structural representation. We call this model as Multi-inducing Sparse Variational Gaussian Process (MultiSVGP). 


\paragraf{Discussion} It is obvious that the conditional distribution in Equation~\ref{eq:decompose_conditional} is not equivalent to the conditional distribution of SVGP in Equation~\ref{eq:naive_model} since the inverse of the matrix sum is not equal to the sum of inverse matrices. Our proposed conditional distribution treats each kernel independently with separate inducing points while the condition in SVGP contains correlations between the kernels which are often complicated under the matrix inverse operator.

\begin{figure*}[h]
    \centering
    \includegraphics[width=0.75\textwidth]{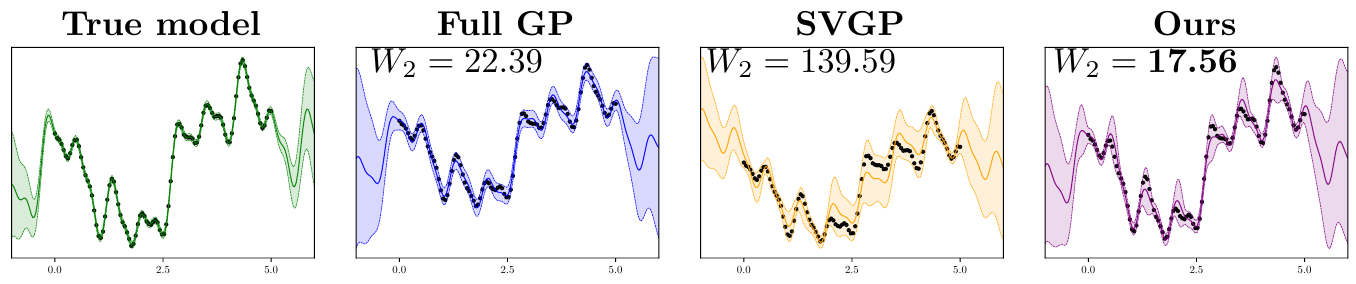}
    \caption{Comparison between the posterior of models. Here, $W_2$ measures the Wasserstein-2 distance between a model and the true model. The posterior obtained from our approach is close to the true model as well as the full GP model. SVGP model struggles to fit the data.}
    \label{fig:comparing}
\end{figure*}

\paragraf{Better fit}  We postulate that the combination of individual conditional Gaussian is still able to learn from data well comparing to other GP models. We justify this with a small experiment in which data are generated from a true model with kernel function $\kSE_1 + \kSE_2 + \kPer_1$. We then fit the data with full GP, SVGP model and our proposed approach. Figure~\ref{fig:comparing} shows the posteriors corresponding to these models along with their Wasserstein-2 distance to the true model. We can observe the high quality of the posterior obtained from our assumption. 

\paragraf{Connection to inter-domain variational Gaussian Process} 
Inter-domain Gaussian process~\cite{inter_domain} helps finding the representative features which lie in difference domains. This work adopts the similar methodology to explain the proposed model.

Consider $m$ Gaussian processes which are governed by different kernels:
\begin{equation}
    g_i(\vx) \sim \mathcal{GP}(0, k_i({\vx}, {\vx}')), \quad i = 1\dots m.
\end{equation}
When applying a linear transformation over a GP, we can obtain a new GP~\cite{inter_domain}. We consider the transformation 
\begin{equation*}
    u_i(\cdot) = \int \phi_i(\vx, \cdot) g_i(\vx) d\vx.
\end{equation*}
The choice of $\phi_i(\vx, {\vz})$ is a Dirac function with the in formation of which inducing point group $\tilde{\vz}$ is in:
\begin{equation*}
    \phi_i(\vx, {\vz}) =\mathbb{I}\{{\vz} \in \mathcal{Z}_i\} \delta({\vz} - \vx).
\end{equation*}
Choosing Dirac delta function $\delta(\cdot)$ is similar to traditional sparse Gaussian process. Whereas $\mathbb{I}\{{\vz} \in \mathcal{Z}_i\}$ provides the membership information of inducing points in the group. If $\vz \in \mathcal{Z}_i$, then $\mathbb{I}(\vz \in \mathcal{Z}_i) = 1$. We combine all $g_i(\vx)$ to get the model in Equation~\ref{eq:full_gp_horseshoe}:
\begin{equation*}
    f(\vx) = \sum_{i=1}^m w_i g_i(\vx) \sim \mathcal{GP}\left(0, \sum_{i=1}^m w_i^2 k_i(\vx, \vx')\right). 
\end{equation*}
Followed by~\cite{inter_domain}, the corresponding (cross-) covariance between $\vU$ and $\vf$ can be obtained as 
\begin{equation}
    \begin{aligned}
    k_{\vu\vf}({\vz},\vx) &= \sum_{i=1}^m w_i \mathbb{I} \{{\vz} \in \mathcal{Z}_i\} k_i({\vz},\vx),\\
    k_{\vu\vu}({\vz}, {\vz'}) &= \sum_{i=1}^m \mathbb{I}\{{\vz} \in \mathcal{Z}_i\} 
    \mathbb{I}\{{\vz}' \in \mathcal{Z}_i\} k_i(\vz, {\vz'}).
    \end{aligned}
    \label{eq:inter_domain_eq}
\end{equation}
From the posterior mean and covariance in Equation~\ref{eq:mean_var_sgp}, we obtain the exact the formula in Equation~\ref{eq:decompose_conditional}.

One may concern about the approximate quality of {MultiSVGP} comparing to SVGP. Here, we argue that our approximate posterior can at least as good as SVGP. Let $\hat{P}$ be the true posterior of GP, $Q_{\multi}$ be our variational approximation in Equation~\ref{eq:decompose_conditional} and $Q_{\single}$ be the variational approximation of SVGP in Equation~\ref{eq:naive_model}. 

We only consider the case\footnote{without loss of generality, $w_i$ is set to $1$} $k(\vx, \vx') = k_1(\vx, \vx') + k_2(\vx, \vx')$. Let $\lambda_i, \lambda_i^{(1)}$, and $\lambda_i^{(2)}$ be the $i$-th operator eigenvalue w.r.t. $k, k_1$, and $k_2$. SVGP has $M$ inducing points. MultiSVGP also has $M$ inducing points in each inducing group. 
According to~\cite{rate_convergence_burt19a}, the bound for the SVGP is
\begin{equation*}
    \KL(Q_{\single}||\hat{P}) \leq \frac{C_{{\single}}}{2\sigma^2_n\delta}\left(1 + \frac{||\vy||^2_2}{\sigma^2_n}\right),
\end{equation*}
 with probability at least $1 - \delta$. Here, $C_{\single} = \sum_{i=M+1}^\infty \lambda_i$, and $\sigma^2_n$ is a Gaussian noise variance. In the case of MultiSVGP, we bound the KL divergence by the following Proposition.

\begin{prop}
    Given $k(\vx, \vx') = k_1(\vx, \vx') + k_2(\vx, \vx')$, with probability at least $1 - \delta$, we have
    \begin{equation*}
        \KL(Q_{\multi}||\hat{P}) \leq \frac{C_{{\multi}}}{2\sigma^2_n\delta}\left(1 + \frac{||\vy||^2_2}{\sigma^2_n}\right),
    \end{equation*}
    with ${C_{\multi}} {=} {N \sum_{i=1}^M\left(\lambda_i - \lambda_i^{(1)} - \lambda_i^{(2)}\right)} {+} { N\sum_{j=M+1}^\infty \lambda_j}$. 
    
    Furthermore, it is true that  
    $C_{\multi} \leq C_{\single}$, making the upper bound of $\KL(Q_{\multi}|| \hat{P})$ is smaller or equal than the upper bound of $\KL(Q_{\single}||\hat{P})$.
    \label{prop:theoretical}
\end{prop}

\begin{proof}
We base on the result in~\cite{rate_convergence_burt19a} on the upper bound of the KL divergence and Equation~\ref{eq:inter_domain_eq} to derive $\textrm{KL}(Q_{\text{multi}}||\hat{P})$. We then use the property of the eigenvalues of sum matrices~\cite{sum_egein, terrance_tao} where $\sum_{i=1}^M\lambda_i \leq \sum_{i=1}^ M\lambda_i^{(1)} + \lambda_i^{(2)}$, to obtain the conclusion. The complete proof is placed in Appendix~\ref{appendix:proof}.
\end{proof}

This is considered a theoretical justification for the comparison in Figure~\ref{fig:comparing}.

\section{Variational inference with shrinkage prior}
With introducing the sparse vector $\vw$ in the kernel selection problem, we tackle the inference problem for this model with variational inference. That is, we consider the variational distribution which will be in the form of factorization between the approximate posterior distribution of GP latent variables and that of sparse vector $\vw$. Let $q(\vw)$ be the variational distribution over $\vw$. The distribution ${q(\vf, \vU, \vw)} {=}{q(\vf, \vU)q(\vw)}$ approximates the true posterior. Following the approximate posterior construction of SVGP, $q(\vf, \vU)$  is still in the similar form, $p(\vf|\vU) q(\vU)$, with ${q(\vU)} {=} {\prod_{i=1}^m q(\vu_i)}$ and $q(\vu_i)$ is parameterized by $\Normal(\vm_i, \vS_i)$. We maximize the evidence lower bound (ELBO)
\begin{equation}
    \begin{aligned}
    \mathcal{L} =& \expect\left[\log \frac{p(\vy, \vf, \vU, \vw)}{q(\vf, \vu, \vw)}\right]\\
      = & \expect_{p(f(\cdot))}\left[\expect_{q(\vw)}[\log p(\vy|\vf, \vw )]\right] \\ & - \textrm{KL}(q(\vU)||p(\vU)) 
      - \textrm{KL}(q(\vw)||p(\vw)).
\end{aligned}
\label{eq:elbo}
\end{equation}
Note that $p(f(\cdot))$ is obtained after marginalizing all $\vu_i$, can be done in the same manner with SVGP. The KL divergence w.r.t. $\vU$ is the sum of the individual KL divergence w.r.t. $\vu_i$.

We describe the subroutine for varitational inference w.r.t. $\vw$, represented by Horseshoe variables, $\tau$ and $\vlambda$. Due to the flat-tailed property of Half-Cauchy distribution, it is reparameterized by double inverse Gamma distributions~\cite{wand2011, horseshoe_bnn}. That is, if $a \sim \mathcal{C}^+(b)$, this is equivalent to $a^2 \sim \ig(\nicefrac{1}{2}, \phi_a^{-1})$ and $\phi_a \sim \ig(\nicefrac{1}{2}, b^{-1})$. Now, we reintroduce the prior containing variables $\{\tau, \vlambda, \phi_{\tau}, \boldsymbol{\phi}_{\vlambda}\}$ as
\begin{align*}
    \tau^2|\phi_\tau &\sim \ig(\nicefrac{1}{2}, \phi_\tau^{-1}), & \phi_\tau & \sim \ig(\nicefrac{1}{2}, A^{-1}),\\
    \lambda_i^2| \phi_{\lambda_i} & \sim \ig(\nicefrac{1}{2}, \phi_{\lambda_i}^{-1}),  & \phi_{\lambda_i} & \sim \ig(\nicefrac{1}{2}, B^{-1}).
\end{align*}
We use the mean-field approach in which the variational distribution $q(\tau, \vlambda, \phi_{\tau}, \boldsymbol{\phi}_{\vlambda})$ is further factorized. Specifically, the variational distributions of $\tau$ and $\lambda_i$ are chosen as log-normal distributions 
\begin{align*}
    q(\tau^2) &= \lognormal(\tau^2;\mu_\tau, \sigma_\tau^2),\\
    q(\lambda_i^2) &= \lognormal(\lambda_i^2; \mu_{\lambda_i},\sigma_{\lambda_i}^2). 
\end{align*}
Whereas $q(\phi_\tau)$ and $q(\phi_{\lambda_i})$ remain inverse Gamma distribution.

As we exchange $\vw$ to $\{\tau,\vlambda\}$, we have the expectation $\expect_{q(\tau)q(\vlambda)}[\log p(\vy|\vf, \tau, \vlambda)]$ which is estimated by Monte Carlo integration.  $q(\tau)$ and $q(\vlambda)$ using reparameterization tricks for Log normal distributions~\cite{reparam_tricks_vae}. 
In particular, since the product $\tau^2\lambda_i^2$ is also Log normal, it can be reparameterized by $\exp(\mu_{\tau}+\mu_{\lambda_i} + \varepsilon (\sigma_\tau + \sigma_{\lambda_i}))$ with $\varepsilon \sim \Normal(0,1)$.
We provide the detailed derivation of ELBO in Appendix~\ref{appendix:variational_horseshoe}.

\paragraph{Computational complexity} Compared to SVGP using single inducing points, MultiSVGP takes $\mathcal{O}(m \max_i\{M_i^2\} b)$ at each optimization iteration with minibatch size $b$. Again, $M_i$ is the number of inducing points $\vu_i$. 

\section{Experimental Evaluations}
\begin{table*}[]
\centering
\caption{Extrapolation performance in UCI benchmarks. Results are aggregated from $10$ independent runs.}
\label{tab:uci_regression}
\scalebox{0.85}{
\begin{tabular}{l|cccc|cccc}
        \toprule
         & \multicolumn{4}{c|}{RMSE}         & \multicolumn{4}{c}{Test log-likelihood} \\ 
         & SVGP-SE & No prior &GP-NKN &  Ours &  SVGP-SE &  No prior  &  GP-NKN   & Ours   \\ \hline
boston   &  $7.30_{\pm 0.21}$         & $7.24_{\pm 0.27}$         &  $5.53_{\pm 0.49}$    &  $\mathbf{5.41}_{\pm 0.10}$         &  $-3.72_{\pm0.07}$ &   $-3.72_{\pm 0.10}$     &       $-3.77_{\pm 0.26}$     &        $\mathbf{-3.24}_{\pm 0.11}$\\
concrete & $9.64_{\pm 0.14}$         &      $8.70_{\pm_{1.05}}$    &   $\mathbf{6.44}_{\pm 0.19}$    &    $7.39_{\pm 0.42}$        &   $-3.54_{\pm 0.01}$       &    $-3.45_{\pm 0.08}$     &       $-\mathbf{3.10}_{\pm 0.01}$     &        $-3.33_{\pm 0.06}$\\
energy   &  $0.83_{\pm 0.07}$         &   $0.69_{\pm 0.18}$    &    $0.41_{\pm 0.03}$      &   $\mathbf{0.37}_{\pm 0.05}$   & $-1.11_{\pm 0.03}$          &   $-1.07_{\pm 0.08}$      &   $\mathbf{-0.54}_{\pm 0.04}$         &  $-0.76_{\pm 0.05}$\\
kin8nm   & $0.11_{\pm 0.00}$     &    $0.11_{\pm 0.08}$   &  $\mathbf{0.09}_{\pm 0.00}$        &    $\mathbf{0.09}_{\pm 0.01}$  &  $0.71_{\pm 0.01}$        &   $0.74_{\pm 0.02}$      &      $\mathbf{1.02}_{\pm 0.05}$      &      $0.89_{\pm 0.01}$  \\
wine     &  $\mathbf{0.62}_{\pm 0.00}$    &  $\mathbf{0.62}_{\pm 0.01}$     &  $0.67_{\pm 0.01}$        &   $\mathbf{0.63}_{\pm 0.01}$   & $-1.04_{\pm 0.00}$         &   $-1.04_{\pm 0.00}$      &      $\mathbf{-1.01}_{\pm 0.01}$      &      $-1.04_{\pm 0.01}$  \\
yacht    &  $1.45_{\pm 0.10}$    &   $1.22_{\pm 0.44}$    &   $0.46_{\pm 0.05}$       &   $\mathbf{0.36}_{\pm 0.05}$   & $-1.91_{\pm 0.14}$    &     $-1.67_{\pm 0.46}$    &     $\mathbf{-0.63}_{\pm 0.02}$       &     $-0.83_{\pm 0.12}$  \\\bottomrule
\end{tabular}
}
\end{table*}
In this section, we first set up choices for compositional kernels. We then justify how the Horseshoe assumption for kernel selection on synthetic data as well as time series data. Finally, we validate our model with regression and classification tasks. Our model is developed based on~\cite{GPflow2017}.
\subsection{Kernel function pool}
\begin{figure}[t]
    \centering
    \includegraphics[width=0.4\textwidth]{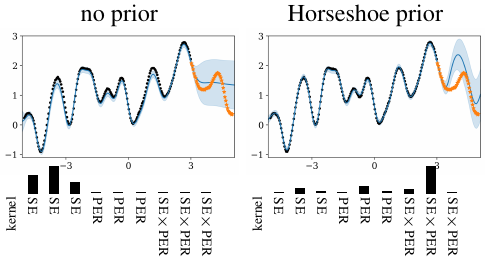}
    \caption{Behavior of Horseshoe prior in kernel selection. Both models predicts the test data (\textcolor{orange}{$\star$}). The bar plots are the weights $w_i$ corresponding to $k_i$.}
    \label{fig:justify}
\end{figure}

\begin{figure}[t]
    \centering
    \input{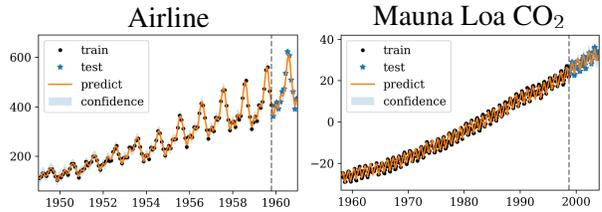}
    \caption{Extrapolation on time series data sets.}
    \label{fig:extrapolate}
\end{figure}

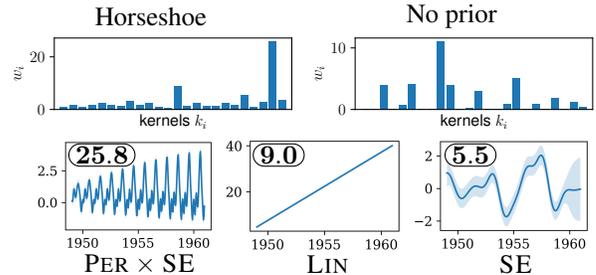
\begin{figure}[t]
    \centering
    \begin{tikzpicture}
    
    \node[inner sep=0pt] at (0.35, 2.4) {Horseshoe};
    \node[inner sep=0pt] at (0.35,1.5) {\includegraphics[width=4cm]{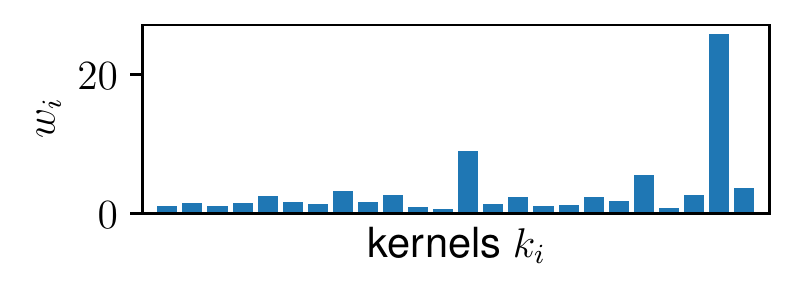}};
    \node[inner sep=0pt] at (4.35, 2.4) {No prior};
    \node[inner sep=0pt] at (4.35,1.5) {\includegraphics[width=4cm]{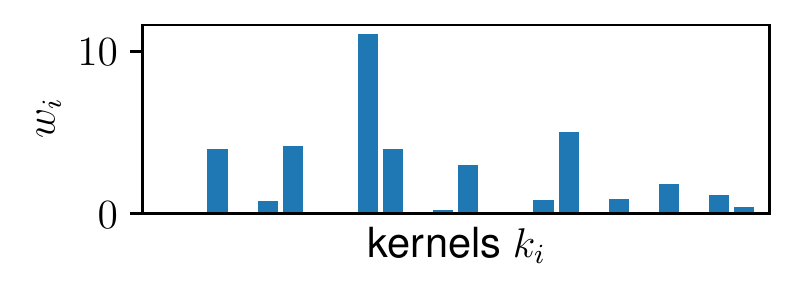}};

    \node[inner sep=0pt] at (0,0) {\includegraphics[width=2.5cm]{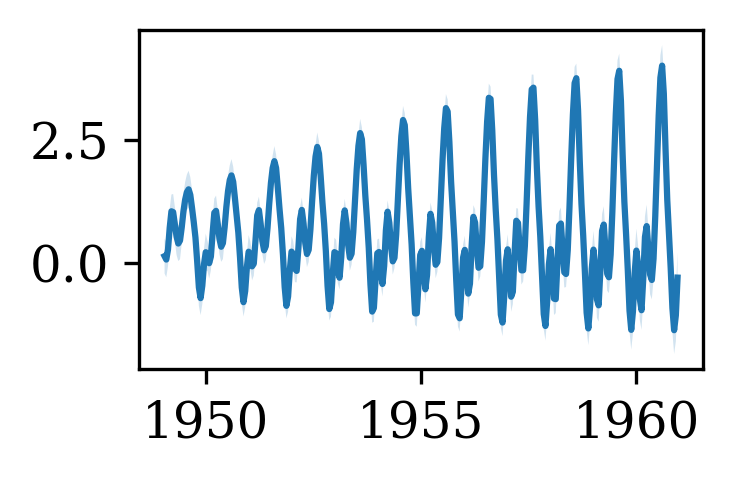}};
    \node[draw, rounded rectangle, inner sep=1pt] at (-0.3, 0.55) {$\mathbf{25.8}$}; 
    \node[inner sep=0pt] at (0.2, -0.9) {$\kPer \times \kSE$};
    
    \node[inner sep=0pt] at (2.5,0) {\includegraphics[width=2.5cm]{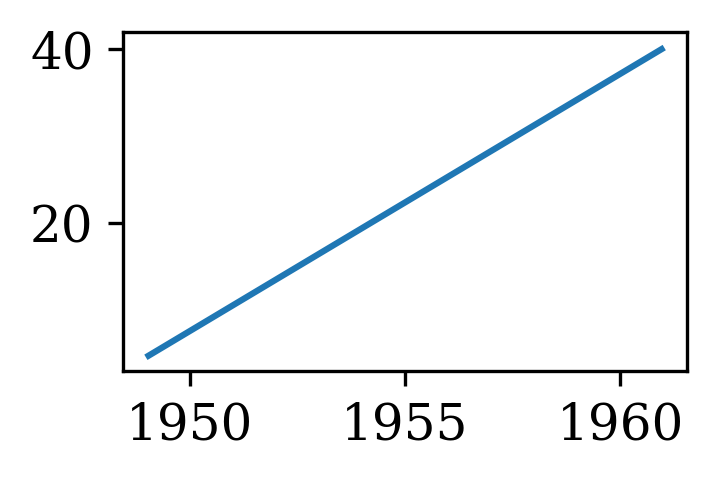}};
    \node[draw, rounded rectangle, inner sep=1pt] at (2.05, 0.55) {$\mathbf{9.0}$}; 
    \node[inner sep=0pt] at (2.7, -0.9) {$\kLin$};
    
    \node[inner sep=0pt] at (5,0) {\includegraphics[width=2.5cm]{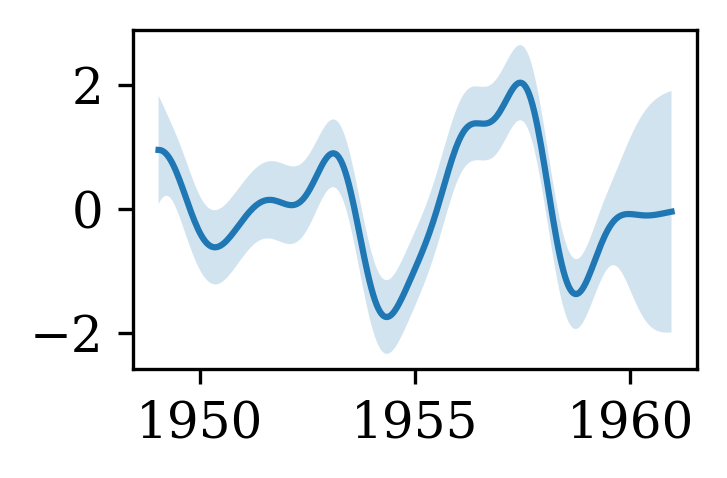}};
    \node[draw, rounded rectangle, inner sep=1pt] at (4.6, 0.55) {$\mathbf{5.5}$}; 
    \node[inner sep=0pt] at (5.2, -0.9) {$\kSE$};
    
    
    \end{tikzpicture}
    \caption{First row: the weights $w_i$ in two cases. Second row: our kernel decomposition for airline data with three most significant components $\mathcal{GP}(\mu_i(\cdot), \Sigma_i(\cdot, \cdot))$. The weights $w_i$ are showed at the upper-left corners.}
    \label{fig:decomposition}
\end{figure}

Now, we present our approach in designing kernel structures for $\{k_i(\vx, \vx')\}$. A kernel function is constructed as a form of multiplicative kernel $\prod_{i=1}^\alpha \mathcal{B}_i$ where $\mathcal{B}_i$ is a base kernel taking from $\kSE, \kLin, \kPer$. In our experiment, the kernel pool is composed of all possible kernels having the multiplicative order up to $2$. We allow duplication in kernels structure. The total number of kernels in the pool is $24$. Each kernel in the pool remains interpretable and can be described by natural language explanations~\cite{Lloyd2014ABCD}. \\
\paragraf{Hyperparameter initialization}~\cite{gpstuff, scale_abcd_kim2018a} suggests two types of hyperparameter initialization: weak prior and strong prior. Unlike existing approaches requiring multiple restarts, we made sure our kernel pool covers both of them.\\
\paragraf{Behavior of Horseshoe prior} 
To see how Horseshoe prior behaves for kernel selection problem, we created a synthetic data $(x_i, y_i)_{i=1}^{100}$ with $x_i \in [-5,5]$ and $y_i$ generated from kernel $\kPer_1 + \kSE \times \kPer_2$. We train our model and compare to the case where there is no prior on weights $\vw$. Figure~\ref{fig:justify} shows our model spike at the relevant kernel structure ($\kSE\times\kPer$) while the model with no prior mistakenly assign weights for local variations ($\kSE$). 
\\\paragraf{Small-sized 1D regression} We verify our model on small-sized data sets: airline passenger volume, Mauna Loa $\textrm{CO}_2$ concentration. Figure~\ref{fig:extrapolate} shows that our model can fit the data well. In the airline data set, the obtained kernel includes $\kPer \times \kSE$, $\kLin$ and $\kSE$ while~\cite{Lloyd2014ABCD} reports $\kLin + \kPer \times \kSE \times \kLin + \kSE$ and a heteroscedastic noise. Also, in the mauna data set, the model can explain the trend and periodicity in data. Our model can reduce the running time to less than $0.5$ hour comparing to $10-12$ hours like~\cite{DuvLloGroetal13, Lloyd2014ABCD} or $2.5-4$ hours like~\cite{scale_abcd_kim2018a}. Figure~\ref{fig:decomposition} provides the visualization of weights $\vw$ found by our model comparing to the model without imposing any prior. This figure also gives the decomposition from $\mathcal{GP}\left(\sum w_i \mu_i(\cdot), w_i^2 \Sigma(\cdot, \cdot)\right)$ corresponding with the most important components.

\noindent\paragraf{Medium-sized 1D regression} We test our model on GEFCOM data set from the Global Energy Forecasting Competition~\cite{gefcom}. The data set has $N=38,070$ data points containing hourly records of energy load from January 2004 to June 2008. We randomly take 90\% of the data set for training and held out 10\% as test data. We compare our model with SVGP with no prior and SVGP with Softmax~\cite{aaai_kernel_selection_svgp}.    

\begin{figure}[t]
    \centering
    \includegraphics[width=0.45\textwidth]{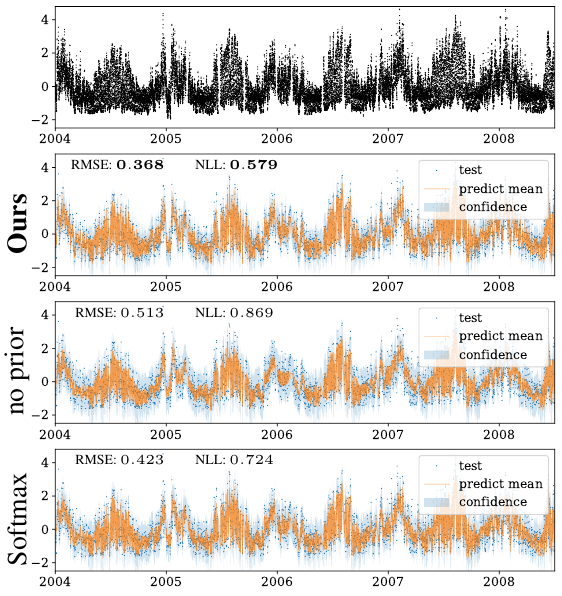}
    \caption{GEFCOM data set. First row is the plot of training data. The next rows are the predictive posterior at test points. Our model outperforms the alternatives in term of root mean square error (RMSE) and test negative log-likelihood (NLL).}
    \label{fig:gefcom}
\end{figure}

Figure~\ref{fig:gefcom} compares the predictive posteriors on the test set. It is clear that our model fits better, giving more accurate predictions as well as uncertainty estimation. The approach in~\cite{aaai_kernel_selection_svgp} takes the second places. The inducing points are associated with complicated kernel function, not divided for each additive kernel. Therefore, the approximate capacity of this model is still more restricted than ours due to Proposition~\ref{prop:theoretical}. 

Our model found $\kSE_1 \times \kPer_1 + \kSE_2 + \kSE_3$ as the kernel structure for this data. This agrees with the kernel function in\cite{lloydgefcom2012} which is manually chosen. Also, our $\kPer$ kernel has periodicity $1.001$ days which also can describe the property that there are peaks in the morning and evening. This is aligned with the result reported in~\cite{scale_abcd_kim2018a}.

\paragraf{Higher-dimension regression} We conducted experiments on UCI data sets~\cite{uci_dataset} including boston, concrete, energy, kin8nm, wine and yatch (see Appendix~\ref{appendix:experiment} for detailed descriptions). We consider baseline models: GP-NKN~\cite{differential_ckl}, SVGP with no shrinkage prior over $\vw$ (no prior), and SVGP with $\kSE$ kernel (SVGP-SE). To justify the extrapolation performance, we projected data onto the principal component of data and sorted data according to the projection~\cite{differential_ckl}. From sorted indices, test data is taken from top $\nicefrac{1}{15}$ and bottom $\nicefrac{1}{15}$ of the data, the remaining is train data. We measure the root mean square error (RMSE) and test log-likelihood in each model.

Table~\ref{tab:uci_regression} shows that our model has a competitive extrapolation capability comparing to GP-NKN. Roughly, our model has better performance in terms of RMSE for most of data sets, except concrete data set. In boston data set, our model performs well for the predictive log-likelihood. Still, GP-NKN consistently outperforms others in this measurement. This is because this model is still considered as a full GP model retaining good uncertainty quantification while the remaining methods including ours are sparse GPs. However, GP-NKN takes significantly more time to train, e.g. in kin8nm. Although the model with no prior has a highly complex kernel, it fails to this extrapolation task. On the other hand, our model with shrinkage prior demonstrates the effect of regularization in kernel selection, resulting in better predictions.

\paragraf{Improving additive GPs} \cite{additive_gp} propose additive kernels for GPs to prevent the local property of kernel functions taking all input dimensions~\cite{local_kernel_machine}. The additive kernel is the sum of lower-dimensional kernel functions which depend on a subset of input variables.

Suppose $D$ is the dimension of data. Let $S_D = \{1,\dots,D\}$ be the index set of dimensions. We adopt this approach and consider the $d$-order additive kernel
\begin{equation*}
    k_d(\vx, \vx') = \sum_{\{i_1, \dots, i_d\} \subseteq S_D}w^2_{i_1\dots i_d}\prod_{i \in \{i_1, \dots, i_d\}} k(\vx[i], \vx'[i]).
    \label{eq:our_additive_kernel}
\end{equation*}
Unlike~\cite{additive_gp} treating weights $w_{i_1\dots i_d}$ equally in the same order $d$, we learn $\vw = [w_{i_1\dots i_d}]$ with our model. We conduct the experiment in three data sets: heart, liver, pima\footnote{taken from~\url{https://github.com/duvenaud/additive-gps}}~\cite{additive_gp} for classification task. The kernel type used here is $\kSE$ kernel. The data sets are randomly split into training ($90\%$ of data) and test ($10\%$ of data) sets.  We first run the model in~\cite{additive_gp} to obtain the most important order $d$. From $d$, we proceed learning $w_{i_1\dots i_d}$. One limitation is that this setting is not scalable w.r.t $D$ as the number of kernels, ${D \choose d}$, increases exponentially. Table~\ref{tab:classification} shows that our model can improve the accuracy of additive GPs by selecting appropriate kernels. On the other hand, the model without any prior even hurts the prediction. In the previous regression task, our model performs poorly in concrete data set since the $1$-order additive kernels is the best fit for this data according to~\cite{additive_gp}. We retrained the model and obtained an improved result with $6.90_{\pm 0.05}$ in RMSE, pushing the result closer to that of GP-NKN.


\begin{table}[]
\centering
\caption{Classification error (in \%) on three data sets.}
\label{tab:classification}
\scalebox{0.85}{
\begin{tabular}{l|cccc}
      & Additive GPs & No prior & Our model  & \#kernels ${D \choose d}$\\ \hline
Heart &    $18.15_{\pm 4.56}$  &    $16.00_{\pm 1.41}$      &  $\mathbf{14.00}_{\pm 2.11}$ &   ${13\choose 1}=13$     \\[1ex] 
Liver &   $30.36_{\pm 8.37}$ &   $40.29_{\pm 6.93}$       &  $\mathbf{27.43}_{\pm 2.52}$ & ${6 \choose 3}=20$\\[1ex] 
Pima  &     $23.99_{\pm 3.46}$     &    $29.87_{\pm 3.72}$      &  $\mathbf{20.52}_{\pm 1.65}$   &  ${8 \choose 6}=28$ 
\end{tabular}
}
\end{table}

We provide an additional experiment comparing our model with the implicit kernel learning~\cite{bbq} in Appendix~\ref{appendix:implicit_kernel_learning}.

\section{Conclusion}
This paper presents a new approximate sparse GP targeting to improve compositional kernel learning for large-scaled data. Moreover, the paper presents a probabilistic kernel selection methods, showing satisfactory results in explaining time series as well as competitive extrapolation performance.

\section*{Acknowledgements}
This work was supported by Institute of Information \& communications Technology Planning \& Evaluation (IITP) grant funded by the Korea government (MSIT)  
(No.2019-0-00075, Artificial Intelligence Graduate School Program (KAIST)). 
This work was partly done at VinAI Research.
We would like to thank Juhyeong Kim for helping revising the paper, and anonymous reviewers for insightful feedback. 
\bibliography{ref}

\newpage
\appendix
\onecolumn
\section{Kernels}
\label{appendix:kernels}
\begin{table}[h]
    \centering
    \begin{tabular}{c|c}
        Name &  Kernel function $k(\vx, \vx')$\\ \midrule
        $\kLin$ & $(\vx - {\ell})^\top(\vx' - {\ell})$\\ [2ex]
        $\kSE$  & $\exp(\frac{||\vx - \vx'||^2}{2{\ell}^2})$ \\ [2ex]
        $\kPer$ & $\exp(-\frac{2\sin^2(\pi ||\vx-\vx'|| /p)}{\ell^2})$
    \end{tabular}
    \caption{Base kernel functions. Note that we do not include the variance hyperparameters in these kernels since it is replaced by $w_i$.}
    \label{tab:my_label}
\end{table}

\section{Proof for Proposition~\ref{prop:theoretical}}
\label{appendix:proof}
\paragraph{Inter-domain covariance}
Before going to the proof of Proposition~\ref{prop:theoretical}, we elucidate the derivation for Equation~\ref{eq:inter_domain_eq}. We can represent our group inducing points as
$$u(\vz) = \sum_{i=1}^m \int \mathbb{I}(\vz \in \mathcal{Z}_i) \delta(\vz - \vx) g_i(\vx) d\vx.$$
Note that $\mathbb{I}(\vz \in \mathcal{Z}_i)$ indicates the membership of inducing points in the group. If $\vz \in \mathcal{Z}_i$, then $\mathbb{I}(\vz \in \mathcal{Z}_i) = 1$. Otherwise, $\mathbb{I}(\vz \in \mathcal{Z}_i) = 0$. The above representation means that only one member in the the group is selected with $\mathbb{I}(\vz \in \mathcal{Z}_i) = 1$.

Following~\cite{inter_domain}, we have the cross-covariance between $u(\vz)$ and $f(\vx)$ as

\begin{align*}
    k_{\vu\vf}(\vz, \vx) = & \expect[u(\vz) f(\vx)] && \\
    = & \expect\left[\sum_{i=1}^m\int \mathbb{I}(\vz \in \mathcal{Z}_i) \delta(\vz - \vx') g_i(\vx') d\vx' \sum_{j=1}^m w_j g_j(\vx) \right] && (\text{by definition of }u \text{ and } f)\\
    = & \sum_{i=1}^m \sum_{j=1}^m w_j \int \mathbb{I}(\vz \in \mathcal{Z}_i) \delta(\vz - \vx')\expect[g_i(\vx') g_j(\vx)] d\vx' && \text{(swapping expectation inside sum and integration)}\\
    = &\sum_{i=1}^m w_i \mathbb{I}(\vz \in \mathcal{Z}_i) \int \delta(\vz - \vx') \expect[g_i(\vx') g_i(\vx)]d\vx' && \text{(since } \Cov(g_i(\vx'), g_j(\vx)) = 0 \text{ with } i \neq j) \\
    = & \sum_{i=1}^m w_i \mathbb{I}(\vz \in \mathcal{Z}_i) \int \delta(\vz - \vx') k_i(\vx', \vx)d\vx' && \text{(since } k_i(\vx', \vx') = \expect[g_i(\vx')g_i(\vx)]) \\
    = & \sum_{i=1}^m w_i \mathbb{I}(\vz \in \mathcal{Z}_i) k_i(\vz, \vx). && \text{(integral with Diract delta function)}
\end{align*}

With similar steps, the covariance between $u$ is obtained as
\begin{align*}
    k_{\vu\vu}(\vz, \vz') 
    = & \expect[u(\vz)u(\vz')] \\
    = & \expect\left[\sum_{i=1}^m\int \mathbb{I}(\vz \in \mathcal{Z}_i) \delta(\vz - \vx) g_i(\vx) d\vx \sum_{j=1}^m \int \mathbb{I}(\vz' \in \mathcal{Z}_j) \delta(\vz' - \vx') g_j(\vx') d\vx' \right] \\
    = & \sum_{i=1}^m \sum_{j=1}^m \mathbb{I}(\vz \in \mathcal{Z}_i) \mathbb{I}(\vz' \in \mathcal{Z}_j) \int \int  \delta(\vz - \vx) \delta(\vz' - \vx') \expect[g_i(\vx)g_j(\vx')] d\vx d\vx' \\
    = & \sum_{i=1}^m \mathbb{I}(\vz \in \mathcal{Z}_i) \mathbb{I}(\vz' \in \mathcal{Z}_i)  \int \int  \delta(\vz - \vx) \delta(\vz' - \vx') \expect[g_i(\vx)g_i(\vx')] d\vx d\vx' \\
    = & \sum_{i=1}^m \mathbb{I}(\vz \in \mathcal{Z}_i) \mathbb{I}(\vz' \in \mathcal{Z}_i)  \int \int  \delta(\vz - \vx) \delta(\vz' - \vx') k_i(\vx, \vx') d\vx d\vx'\\
    = & \sum_{i=1}^m \mathbb{I}(\vz \in \mathcal{Z}_i) \mathbb{I}(\vz' \in \mathcal{Z}_i)  k(\vz, \vz').
\end{align*}

\paragraph{Proof of Proposition~\ref{prop:theoretical}}
In Lemma 1 of~\cite{rate_convergence_burt19a}, we have
\begin{equation*}
    \KL(Q_{\multi}||\hat{P}) \leq \frac{t}{2\sigma_n^2} \left (1 + \frac{||\vy||^2_2}{\sigma^2_n + t}\right).
\end{equation*}

Here $t = \textrm{trace}(\vK_{\vf\vf} - \underbrace{\vK_{\vu\vf}^\top \vK_{\vu\vu}^{-1} \vK_{\vu\vf}}_{\mathbf{Q}_{\vf\vf}})$. $\vK_{\vu\vf}$ is the cross-covariance matrix between inducing points $\vU$ and $\vf$. $\vK_{\vu\vu}$ is the covariance matrix of $\vU$.

Recall the covariance between $\vU$ and $\vf$ in Equation~\ref{eq:inter_domain_eq}:

\begin{align*}
    k_{\vu\vf}(\vz, \vx) = & \mathbb{I}\{\vz \in \mathcal{Z}_1\} k_1(\vz, \vx) + \mathbb{I}\{\vz \in \mathcal{Z}_2\} k_2(\vz, \vx),\\
    k_{\vu\vu}(\vz, \vz') = & \mathbb{I}\{\vz \in \mathcal{Z}_1\}\mathbb{I}\{\vz' \in \mathcal{Z}_1\}k_1(\vz, \vz') + \mathbb{I}\{\vz \in \mathcal{Z}_2\}\mathbb{I}\{\vz' \in \mathcal{Z}_2\}k_2(\vz, \vz'),
\end{align*}
where $\mathbb{I}(\vz \in \mathcal{Z}_i) = 1$ if $\vz \in \mathcal{Z}_i$, otherwise it is equal $0$. Then, we can compute
\begin{equation*}
    \vK_{\vu\vf} = \Cov \left(\begin{bmatrix}\vu_1\\
    \vu_2\end{bmatrix}, \vf\right) = \begin{bmatrix} \vK_{\vu_1\vf}\\
    \vK_{\vu_2\vf}\end{bmatrix}, 
\end{equation*}

\begin{equation*}
    \vK_{\vu\vu} = \Cov \left(\begin{bmatrix}\vu_1\\
    \vu_2\end{bmatrix}, \begin{bmatrix}\vu_1\\
    \vu_2\end{bmatrix}\right) = \begin{bmatrix} \vK_{\vu_1\vu_1} & \mathbf{0}\\
    \mathbf{0} & \vK_{\vu_2\vu_2}\end{bmatrix},
\end{equation*}
where $\vK_{\vu_1\vf}$ is the covariance matrix computed from $k_1(\cdot, \cdot)$ with $\vu_1 \in \mathcal{Z}_1$, and $K_{\vu_1\vu_1}$ is the covariance matrix of $\vu_1 \in \mathcal{Z}_1$ computed from $k_1(\cdot, \cdot)$. The same is applied to $\vK_{\vu_2\vf}$ and $\vK_{\vu_2\vu_2}$.

From $\mathbf{Q}_{\vf\vf} = \vK_{\vu\vf}^\top \vK_{\vu\vu}^{-1}\vK_{\vu\vf} $ and block matrix multiplication, we can obtain 
$$\mathbf{Q}_{\vf\vf}=  \vK_{\vu_1\vf}^\top\vK_{\vu_1\vu_1}^{-1}\vK_{\vu_1\vf} + \vK_{\vu_2\vf}^\top\vK_{\vu_1\vu_2}^{-1}\vK_{\vu_2\vf}.$$

Let $\psi^{(1)}_i$ and $\psi^{(2)}_i$  be the eigenfunctions of the covariance operators w.r.t. $k_1(\vx, \vx')$ and $k_2(\vx, \vx')$. Following the similar steps in~\cite{rate_convergence_burt19a}, we obtain the individual terms in $\mathbf{Q}_{\vf\vf}$  under the eigenfeature representation as
    $$[\vK_{\vf\vu_1}\vK_{\vu_1\vu_1}^{-1}\vK_{\vu_1\vf}]_{c,r} = \sum_{i=1}^M \lambda_i^{(1)} \psi^{(1)}_i(\vx_c)\psi^{(1)}_i(\vx_r),$$
    $$[\vK_{\vf\vu_2}\vK_{\vu_2\vu_2}^{-1}\vK_{\vu_2\vf}]_{c,r} = \sum_{i=1}^M \lambda_i^{(2)} \psi^{(2)}_i(\vx_c)\psi^{(2)}_i(\vx_r).$$

By Mercer's theorem and with $\psi_i(\vx)$ be the eigenfunctions of covariance operator for $k(\vx,\vx')$, we have
    $$k(\vx, \vx') = \sum_{i=1}^{\infty} \lambda_i \psi_i(\vx)\psi_i(\vx').$$

Then the entry at $(n,n)$ of $\vK_{\vf\vf} - \mathbf{Q}_{\vf\vf}$ is
\begin{equation*}
    [\vK_{\vf\vf} - \mathbf{Q}_{\vf\vf}]_{n,n} = \sum_{i=1}^{\infty} \lambda_i \psi_i^2(\vx_n) - \left(\sum_{i=1}^M \lambda_i^{(1)} (\psi_i^{(1)})^2(\vx_n) + \sum_{i=1}^M \lambda_i^{(2)} (\psi_i^{(2)})^2(\vx_n) \right).
\end{equation*}

From this, we can have the expectation of $t$

\begin{equation*}
    \expect_\vx[t] = N\sum_{i=1}^\infty\lambda_i - N \left(\sum_{i=1}^M \lambda_i^{(1)} + \sum_{i=1}^M \lambda_i^{(2)}\right).
\end{equation*}
Here the eigenfunction terms are disappeared. Because $\expect[\psi^2_i(\vx)] = \int \psi^2(\vx)p(\vx)d\vx = 1$. Similarly, $\expect[(\psi_i^{(1)})^2(\vx)] = \expect[(\psi_i^{(2)})^2(\vx)] = 1$.

According to~\cite{rate_convergence_burt19a}, we apply the Markov's inequality, we have, with probability at least $1 - \delta$, 
\begin{equation*}
    \KL(Q_{\multi}||\hat{P}) \leq \frac{C_{\multi}}{2\sigma^2\delta}\left(1 + \frac{||\vy||^2_2}{\sigma^2_n}\right),
\end{equation*}
where $C_{\multi} = N \sum_{i=1}^M\left(\lambda_i - \lambda_i^{(1)} - \lambda_i^{(2)}\right) + N\sum_{j=M+1}^\infty \lambda_j$.

Comparing to $C_{\single} = \sum_{i=M+1}^{\infty}\lambda_i$, we use the result in~\cite{sum_egein, terrance_tao} where
\begin{equation*}
    \sum_{i=1}^M \lambda_i  \leq \sum_{i=1}^M\lambda_i^{(1)} + \sum_{i=1}^M \lambda_i^{(2)}.
\end{equation*}

We can conclude that the upper bound of $\KL(Q_{\multi}||\hat{P})$ is smaller than the upper bound of $\KL(Q_{\single}||\hat{P})$.

\section{Detail of variational inference}
\label{appendix:variational_horseshoe}
\paragraf{Prior} Recall the prior over $\tau, \vlambda, \phi_\tau, \vphi_{\vlambda}$ after reparameterization is
\begin{align*}
    \tau^2| \phi_\tau & \sim  \ig(\tau^2|\nicefrac{1}{2}, \phi_\tau^{-1}),\\
    \phi_\tau & \sim \ig(\phi_\tau|\nicefrac{1}{2}, A^{-1}),\\
    \lambda_i^2| \phi_{\lambda_i} & \sim \ig(\lambda_i^2|\nicefrac{1}{2}, \phi_{\lambda_i}^{-1}), \quad i = 1\dots m,\\
    \phi_{\lambda_i} & \sim \ig(\phi_{\lambda_i}|\nicefrac{1}{2}, B^{-1}).
\end{align*}

\paragraf{Variational distribution} The variational distributions of $\tau$ and $\lambda_i$ are in the form of log normal distribution. 
\begin{align*}
    q(\tau^2) &= \lognormal(\tau^2|m_\tau, \sigma^2_\tau)\\
    q(\lambda_i^2) &= \lognormal(\lambda_i^2|m_{\lambda_i}, \sigma^2_{\lambda_i}), \quad i = 1\dots m. 
\end{align*}
On the other hand, the variational distributions of the auxiliary variables $\phi_\tau$ and $\phi_{\lambda_i}$ remain as Inverse Gamma distributions
\begin{align*}
    q(\phi_\tau) &= \ig(\phi_\tau|s_\tau, r_\tau)\\
    q(\phi_{\lambda_i}) & = \ig(\phi_{\lambda_i}|s_{\lambda_i}, r_{\lambda_i}), \quad i = 1\dots m. 
\end{align*}
\paragraf{KL divergence} As $\vw = \{\tau, \vlambda, \phi_\tau, \vphi_{\vlambda}\}$, the KL divergence $\KL(q(\vw)||p(\vw))$ becomes
\begin{equation}
    \begin{aligned}
    &\KL\left( q(\tau^2) q(\phi_\tau) \prod_i q(\lambda_i^2) q(\phi_{\lambda_i}) || p(\tau^2|\phi_\tau) p(\phi_\tau) \prod_i p(\lambda_i|\phi_{\lambda_i}) p(\lambda_i)\right)\\
    = & \textcolor{green}{H[q(\tau^2)]} + \textcolor{red}{{H[q(\phi_\tau)]}} + \textcolor{green}{\sum_i H [q(\lambda_i^2)]} + \textcolor{red}{\sum_i H[q(\phi_{\lambda_i})] }+ \\
    & \textcolor{blue}{\expect_{q(\tau^2)q(\phi_\tau)}[\log p(\tau^2|\phi_\tau)]} + \textcolor{red}{\expect_{q(\phi_\tau)}[\log p(\phi_\tau)] }+ \textcolor{blue}{\sum_i \expect_{q(\lambda_i)q(\phi_{\lambda_i})}[\log p(\lambda_i^2|\phi_{\lambda_i})]} + \textcolor{red}{\sum_i \expect_{q(\phi_{\lambda_i})} [\log p(\phi_{\lambda_i})]}.
    \end{aligned}
    \label{eq:kl_tau_lambda}
\end{equation}
where $H[\cdot]$ denotes the entropy of a distribution. 

Individual terms will be explained as following. The entropy terms will be computed as
\textcolor{green}{
\begin{align*}
    H[q(\tau^2)] &= \mu_\tau + \frac{1}{2}\log (2\pi e \sigma_\tau^2),\\
    H[q(\lambda_i^2)] & = \mu_{\lambda_i} + \frac{1}{2}\log (2\pi e \sigma_{\lambda_i}^2).
\end{align*}
}

The expectations of log prior can be derived as

\begin{align*}
    \textcolor{blue}{\expect_{q(\tau^2)q(\phi_\tau)}[\log p(\tau^2|\phi_\tau)]} & = \expect_{q(\tau^2)q(\phi_\tau)}\left[\log \ig(\tau^2| \nicefrac{1}{2}, \phi_\tau^{-1})\right] \\
    & = \expect_{q(\tau^2)q(\phi_\tau)}[-\frac{1}{2}\log \phi_\tau - \log \Gamma(\nicefrac{1}{2}) - \frac{3}{2} \log(\tau^2) - \frac{1}{\tau^2 \phi_\tau}] \\
    & = -\frac{1}{2}\expect_{q(\phi_\tau)}[\log \phi_\tau] - \log \Gamma(\nicefrac{1}{2}) - \frac{3}{2} \expect_{q(\tau^2)}[\log(\tau^2)] - \expect_{q(\tau^2)}[\tau^{-2}]\expect_{\phi_\tau}[\phi_\tau^{-1}],
\end{align*}
where the individual terms can be calculated as
\begin{align*}
    \expect_{q(\phi_\tau)}[\log \phi_\tau] &= \log r_{\tau} - \psi(s_\tau), && {\text{(Inverse Gamma distribution property)}}\\
    \expect_{q(\phi_\tau)}[\phi_\tau^{-1}] & = \frac{s_\tau}{r_{\tau}}, && {\text{(Inverse Gamma distribution property)}} \\
    \expect_{q(\tau^2)}[\log(\tau^2)] & = \mu_\tau && \text{(compute from log normal distribution)}\\
    \expect_{q(\tau^2)}[\tau^{-2}] & = \exp(-\mu_\tau + \frac{1}{2}\sigma_\tau^2). && \text{(Log normal distribution property)}
\end{align*}
Here, $\psi(\cdot)$ is the digamma function (is not the same with $\psi$ in Section~\ref{appendix:proof}). 
Similarly, we can obtain the expectation of log prior w.r.t to $\lambda_i$
\begin{align*}
    \textcolor{blue}{\expect_{q(\lambda_i^2)q(\phi_{\lambda_i})}[\log p(\lambda_i^2|\phi_{\lambda_i})]} = -\frac{1}{2}\expect_{q(\phi_{\lambda_i})}[\log \phi_{\lambda_i}] - \log \Gamma(\nicefrac{1}{2}) - \frac{3}{2} \expect_{q(\lambda_i^2)}[\log(\lambda_i^2)] - \expect_{q(\lambda_i^2)}[\lambda_i^{-2}]\expect_{\phi_{\lambda_i}}[\phi_{\lambda_i}^{-1}]
\end{align*}

We intentionally do not write the explicit form of \textcolor{red}{$H[q(\phi_\tau)]$}, \textcolor{red}{$H[q(\phi_{\lambda_i})]$}, \textcolor{red}{$\expect_{q(\phi_\tau)}[\log p(\phi_\tau)]$} and \textcolor{red}{$\expect_{q(\phi_{\lambda_i})} [\log p(\phi_{\lambda_i})]$} because the variables $\phi_\tau$ and $\vphi_{\vlambda}$ do not follow an optimization but are updated by the following.
\paragraph{Closed-form update for $q(\phi_\tau)$ and $q(\phi_{\lambda_i})$} Under the mean-field assumption on variational variables $\tau, \vlambda, \phi_\tau, \vphi_{\vlambda}$, we can obtain the closed-form optimal solution w.r.t. the auxiliary variables $\phi_\tau, \vphi_{\vlambda}$~\cite{mean_field_shrinkage}. That is, after each optimization step on other variables, we update $q(\phi_\tau)$ and $q(\phi_{\lambda_i})$ by
\begin{equation}
    \begin{aligned}
        q(\phi_\tau) &= \ig(s_\tau = 1, r_\tau = \expect[\tau^{-2}] + A^{-2}),\\
        q(\phi_{\lambda_i}) & = \ig(s_{\lambda_i} = 1, r_{\lambda_i} = \expect[\lambda_i^{-2}] + B^{-2}).
    \end{aligned}
    \label{eq:update_tau_lambda}
\end{equation}

\paragraf{Evidence lower bound} Recap that the evidence lower bound is in the following form:
\begin{equation}
    \begin{aligned}
    \mathcal{L}
      = & \expect_{p(f(\cdot))}\left[\expect_{q(\tau, \vlambda)}[\log p(\vy|\vf, \tau, \vlambda )]\right]  - \textrm{KL}(q(\vU)||p(\vU)) 
      - \textrm{KL}(q(\tau, \vlambda)||p(\tau, \vlambda)).
\end{aligned}
\label{eq:appendix_elbo}
\end{equation}
Note that the expectation w.r.t $\tau, \vlambda$ is estimated by Monte Carlo integration. During training, we draw one sample $\tau_S$ and $\vlambda_S$ by the reparameterization trick for the product $\tau_S\lambda_{i_S} = \exp(\mu_{\tau}+\mu_{\lambda_i} + \varepsilon (\sigma_\tau + \sigma_{\lambda_i}))$ where $\varepsilon \sim \Normal(0, 1)$. 

The following algorithm describes our variational inference

\begin{algorithm}
\caption{Variational inference for MultiSVGP with Horseshoe prior}
    \begin{algorithmic}
    \REQUIRE Data $\mathcal{D}=\{\vX, \vy\}$, a set of kernel function 
    $\{k_i(\vx,\vx')\}_{i=1}^m$
    \STATE Initialize kernel hyperparameters, variational parameters $\{\mu_\tau, \sigma_\tau^2, \mu_{\lambda_i}, \sigma^2_{\lambda_i}, s_\tau, r_\tau, s_{\lambda_i}, r_{\lambda_i}\}$
    \FOR {within a number of iterations}
        \STATE Sample a minibatch $(\vx_b, \vy_b)$
        \STATE Sample $\tau_S, \vlambda_S$ with $\tau_S\lambda_{i_S} = \exp(\mu_{\tau}+\mu_{\lambda_i} + \varepsilon (\sigma_\tau + \sigma_{\lambda_i}))$ where $\varepsilon \sim \Normal(0, 1)$
        \STATE Compute $\expect_{p(f(\vx_b))}\left[\expect_{q(\tau, \vlambda)}[\log p(\vy_b|\vf, \tau, \vlambda )]\right] \approx \expect_{p(f(\vx_b))}\left[\log p(\vy_b|\vf, \tau_S, \vlambda_S)\right]$
        \STATE Compute $\KL(q(\vU)||p(\vU))$ as the sum of $\KL(q(\vu_i)||p(\vu_i))$
        \STATE Compute $\KL(q(\tau, \vlambda)|| p(\tau, \vlambda))$ by Equation~\ref{eq:kl_tau_lambda}
        \STATE Compute ELBO $\mathcal{L}$ based on Equation~\ref{eq:appendix_elbo}
        \STATE Perform an optimization step for ELBO $\mathcal{L}$
        \STATE Update $q(\phi_\tau)$ and $q(\phi_{\lambda_i})$ by Equation~\ref{eq:update_tau_lambda}
    \ENDFOR
    \end{algorithmic}
\end{algorithm}

\section{Experiments}
\label{appendix:experiment}




\subsection{Description of regression data set}
See Table~\ref{tab:desc_regression}
\begin{table}[h]
\centering
\caption{Description of UCI data sets~\cite{uci_dataset}}
\label{tab:desc_regression}
\begin{tabular}{l|c|c|l}
\toprule
Data set & \# data $N$ & Dimension $D$ & Description \\ \hline
boston   &   506      &     13      &       Boston housing price      \\
concrete &  1030       &    8   &      Predict concrete compressive strength       \\ 
energy   &  768       &     8      &     Predict energy efficiency for buildings        \\
kin8nm   &  8192       &    8       &    Kinematics of an 8 link robot arm         \\
wine     &  1599       &    22      &     Wine quality data set        \\
yacht    &    308     &    7       &   Prediction of residuary resistance of sailing yachts         \\\bottomrule
\end{tabular}
\end{table}
\subsection{Description of classification task}
See Table~\ref{tab:desc_classification}
\begin{table}[h]
\centering
\caption{Description of heart, liver, pima data set}
\label{tab:desc_classification}
\begin{tabular}{l|c|c|l}
\toprule
Data set & \# data $N$ & Dimension $D$ & Description \\ \hline
heart   &    303     &     13      &      Predict the presence of heart disease       \\
liver   &     345    &     6      &     Predict liver disorders        \\
pima &    768     &       8    &       Pima Indians Diabetes Database      \\ \bottomrule
\end{tabular}
\end{table}

\subsection{Detailed experiment setup}
\paragraph{Number of inducing points} The number of inducing points used in airline and mauna loa data set is $100$. In the remaining experiments, the number of inducing points is set to $200$. 

\paragraph{Horseshoe hyperparameter} We choose the hyperparameters $A=1, B=1$ in Horseshoe prior. Varying these hyperparameters (between $[1,3]$) does not affect much to the results since the maximum number of kernels is not big, about $24$ (in regression task) to $28$ (in classification task). 

\paragraph{Experiment with softmax assumption~\cite{aaai_kernel_selection_svgp}} We follow the model candidates in~\cite{aaai_kernel_selection_svgp} where there are $12$ possible GP models to selects. We cannot run the setting in which there are $144$ candidate models described in~\cite{aaai_kernel_selection_svgp}. We implement this model by introducing a deterministic softmax weights to select model which is slightly different from~\cite{aaai_kernel_selection_svgp}. Yet, it still reflects the choice of model selection.

Table~\ref{tab:uci_regression_2} includes the results considering the softmax assumption. Our model still outperforms the model with softmax in terms of both RMSE and test log-likelihood.
\begin{table*}[h]
\centering
\caption{Extrapolation performance in UCI benchmarks. Results are aggregated from $10$ independent runs.}
\label{tab:uci_regression_2}
\scalebox{0.8}{
\begin{tabular}{l|ccccc|ccccc}
        \toprule
         & \multicolumn{5}{c|}{RMSE}         & \multicolumn{5}{c}{Test log-likelihood} \\ 
    Data set     & SVGP-SE & No prior & Softmax &GP-NKN &  Ours &  SVGP-SE &  No prior & Sotfmax  &  GP-NKN   & Ours   \\ \hline
boston   &  $7.30_{\pm 0.21}$   & $7.24_{\pm 0.27}$    & $6.90_{\pm 0.34}$&  $5.53_{\pm 0.49}$    &  $\mathbf{5.41}_{\pm 0.10}$         &  $-3.72_{\pm0.07}$ &   $-3.72_{\pm 0.10}$   & $-3.36_{\pm 0.04}$ &       $-3.77_{\pm 0.26}$     &        $\mathbf{-3.24}_{\pm 0.11}$\\
concrete & $9.64_{\pm 0.14}$         &      $8.70_{\pm_{1.05}}$ & $7.46_{\pm 0.43}$ &   $\mathbf{6.44}_{\pm 0.19}$    &    $7.39_{\pm 0.42}$        &   $-3.54_{\pm 0.01}$       &    $-3.45_{\pm 0.08}$     &   $-3.39_{\pm 0.04}$  &  $-\mathbf{3.10}_{\pm 0.01}$     &        $-3.33_{\pm 0.06}$\\
energy   &  $0.83_{\pm 0.07}$         &   $0.69_{\pm 0.18}$    & $0.51_{\pm 0.05}$ &    $0.41_{\pm 0.03}$      &   $\mathbf{0.37}_{\pm 0.05}$   & $-1.11_{\pm 0.03}$          &   $-1.07_{\pm 0.08}$      & $-0.89_{\pm 0.04}$ &   $\mathbf{-0.54}_{\pm 0.04}$         &  $-0.76_{\pm 0.05}$\\
kin8nm   & $0.11_{\pm 0.00}$     &    $0.11_{\pm 0.08}$   & $0.13_{\pm 0.00}$ & $\mathbf{0.09}_{\pm 0.00}$        &    $\mathbf{0.09}_{\pm 0.01}$  &  $0.71_{\pm 0.01}$        &   $0.74_{\pm 0.02}$  & $0.60_{\pm 0.01}$    &      $\mathbf{1.02}_{\pm 0.05}$      &      $0.89_{\pm 0.01}$  \\

wine     &  $\mathbf{0.62}_{\pm 0.00}$    &  $\mathbf{0.62}_{\pm 0.01}$   & $0.64_{\pm 0.01}$  &  $0.67_{\pm 0.01}$        &   $\mathbf{0.63}_{\pm 0.01}$   & $-1.04_{\pm 0.00}$         &   $-1.04_{\pm 0.00}$      &$-1.07_{\pm 0.01}$&     $\mathbf{-1.01}_{\pm 0.01}$      &      $-1.04_{\pm 0.01}$  \\
yacht    &  $1.45_{\pm 0.10}$    &   $1.22_{\pm 0.44}$    & $1.81_{\pm 0.53}$&  $0.46_{\pm 0.05}$       &   $\mathbf{0.36}_{\pm 0.05}$   & $-1.91_{\pm 0.14}$    &     $-1.67_{\pm 0.46}$    &   $-2.16_{\pm 0.13}$   & $\mathbf{-0.63}_{\pm 0.02}$       &     $-0.83_{\pm 0.12}$  \\\bottomrule
\end{tabular}
}
\end{table*}

\subsection{Compare to implicit kernel learning}
\label{appendix:implicit_kernel_learning}
This experiment compares between our proposed method and Black Box Quantiles (BBQ)~\cite{bbq}. Data sets and experimental setups are followed~\cite{bbq}. Table~\ref{tab:compare_implicit} shows the RMSE and MNLL of two methods. Our model shows competitive results in most data sets. On the other hand, in the intra-filling tasks on two data sets (pores and tread) BBQ is better. In these two data sets, our model comes with the second place according to the results of remaining alternative models in~\cite{bbq}. 

\begin{table*}[t]
\centering
\caption{Comparison to an implicit kernel learning approach}
\label{tab:compare_implicit}
\begin{tabular}{l|cc|cc} \toprule
\multirow{2}{*}{Data set} & \multicolumn{2}{c|}{RMSE} & \multicolumn{2}{c}{MNLL} \\ \cline{2-5} 
                          & BBQ         & Ours        & BBQ         & Ours        \\ \hline
CO2                       & $\mathbf{0.068}$      & $0.157$       & $-1.242$      & $\mathbf{-1.649}$      \\
Passenger                 & $0.096$       & $\mathbf{0.035}$       & $-0.610$      & $\mathbf{-1.702}$      \\
Concrete                  & $0.124$       & $\mathbf{0.086}$       & $-0.577$      & $\mathbf{-0.742}$      \\
Noise                     & $0.138$       & $\mathbf{0.074}$       & $-0.173$      & $\mathbf{-0.922}$      \\
Rubber                    & $0.248$       & $\mathbf{0.225}$       & $0.523$       & $\mathbf{-0.010}$      \\
Pores                     & $\mathbf{0.256}$       & $0.574$       & $\mathbf{0.335}$       & $0.898$       \\
Tread                     & $\mathbf{0.114}$       & $0.123$       & $\mathbf{-0.754}$      & $-0.644$     \\
\bottomrule
\end{tabular}
\end{table*}

\end{document}